\documentclass[3p,,preprint,12pt]{elsarticle}
\makeatletter\if@twocolumn\PassOptionsToPackage{switch}{lineno}\else\fi\makeatother

\usepackage{tabulary,xcolor}
\usepackage{amsfonts,amsmath,amssymb}
\usepackage[T1]{fontenc}
\makeatletter
\let\save@ps@pprintTitle\ps@pprintTitle
\def\ps@pprintTitle{\save@ps@pprintTitle\gdef\@oddfoot{\footnotesize\itshape \null\hfill\today}}
\def\hlinewd#1{%
  \noalign{\ifnum0=`}\fi\hrule \@height #1%
  \futurelet\reserved@a\@xhline}
\def\tbltoprule{\hlinewd{.8pt}\\[-12pt]}
\def\tblbottomrule{\noalign{\vspace*{6pt}}\hline\noalign{\vspace*{2pt}}}
\def\tblmidrule{\noalign{\vspace*{6pt}}\hline\noalign{\vspace*{2pt}}}
\AtBeginDocument{\ifNAT@numbers \biboptions{sort&compress}\fi}
\makeatother

\usepackage{ifluatex}
\ifluatex
\usepackage{fontspec}
\defaultfontfeatures{Ligatures=TeX}
\usepackage[]{unicode-math}
\unimathsetup{math-style=TeX}
\else 
\usepackage[utf8]{inputenc}
\fi 
\ifluatex\else\usepackage{stmaryrd}\fi

\usepackage{url,multirow,morefloats,floatflt,cancel,tfrupee}
\makeatletter

\AtBeginDocument{\@ifpackageloaded{textcomp}{}{\usepackage{textcomp}}}
\makeatother
\usepackage{colortbl}
\usepackage{xcolor}
\usepackage{pifont}
\usepackage[nointegrals]{wasysym}
\makeatletter

\def\mcWidth#1{\csname TY@F#1\endcsname+\tabcolsep}

\def\cAlignHack{\rightskip\@flushglue\leftskip\@flushglue\parindent\z@\parfillskip\z@skip}
\def\rAlignHack{\rightskip\z@skip\leftskip\@flushglue \parindent\z@\parfillskip\z@skip}

\@ifundefined{etal}{}{}

\usepackage{ifxetex}
\ifxetex\else\if@twocolumn\@ifpackageloaded{stfloats}{}{\usepackage{dblfloatfix}}\fi\fi

\AtBeginDocument{
\expandafter\ifx\csname eqalign\endcsname\relax
\def\eqalign#1{\null\vcenter{\def\\{\cr}\openup\jot\m@th
  \ialign{\strut$\displaystyle{##}$\hfil&$\displaystyle{{}##}$\hfil
      \crcr#1\crcr}}\,}
\fi
}

\AtBeginDocument{%
  \@ifpackageloaded{endfloat}%
   {\renewcommand\efloat@iwrite[1]{\immediate\expandafter\protected@write\csname efloat@post#1\endcsname{}}}{\newif\ifefloat@tables}%
}%

\def\BreakURLText#1{\@tfor\brk@tempa:=#1\do{\brk@tempa\hskip0pt}}
\let\lt=<
\let\gt=>
\def\processVert{\ifmmode|\else\textbar\fi}

\@ifundefined{subparagraph}{
\def\subparagraph{\@startsection{paragraph}{5}{2\parindent}{0ex plus 0.1ex minus 0.1ex}%
{0ex}{\normalfont\small\itshape}}%
}{}

\newcommand\role[1]{\unskip}
\newcommand\aucollab[1]{\unskip}
  
\@ifundefined{tsGraphicsScaleX}{\gdef\tsGraphicsScaleX{1}}{}
\@ifundefined{tsGraphicsScaleY}{\gdef\tsGraphicsScaleY{.9}}{}
\def\checkGraphicsWidth{\ifdim\Gin@nat@width>\linewidth
	\tsGraphicsScaleX\linewidth\else\Gin@nat@width\fi}

\def\checkGraphicsHeight{\ifdim\Gin@nat@height>.9\textheight
	\tsGraphicsScaleY\textheight\else\Gin@nat@height\fi}

\def\fixFloatSize#1{}%\@ifundefined{processdelayedfloats}{\setbox0=\hbox{\includegraphics{#1}}\ifnum\wd0<\columnwidth\relax\renewenvironment{figure*}{\begin{figure}}{\end{figure}}\fi}{}}
\let\ts@includegraphics\includegraphics

\def\inlinegraphic[#1]#2{{\edef\@tempa{#1}\edef\baseline@shift{\ifx\@tempa\@empty0\else#1\fi}\edef\tempZ{\the\numexpr(\numexpr(\baseline@shift*\f@size/100))}\protect\raisebox{\tempZ pt}{\ts@includegraphics{#2}}}}

\AtBeginDocument{\def\includegraphics{\@ifnextchar[{\ts@includegraphics}{\ts@includegraphics[width=\checkGraphicsWidth,height=\checkGraphicsHeight,keepaspectratio]}}}

\DeclareMathAlphabet{\mathpzc}{OT1}{pzc}{m}{it}

\def\URL#1#2{\@ifundefined{href}{#2}{\href{#1}{#2}}}

\def\UrlOrds{\do\*\do\-\do\~\do\'\do\"\do\-}%
\g@addto@macro{\UrlBreaks}{\UrlOrds}

\edef\fntEncoding{\f@encoding}

\makeatother

\newif\ifmultipleabstract\multipleabstractfalse%
\makeatletter
\def\ps@pprintTitle{\save@ps@pprintTitle\gdef\@oddfoot{\footnotesize\hspace*{.5\textwidth}\thepage\itshape \null\hfill\today}}
\makeatother
          
\usepackage[flushleft]{threeparttablex}

\usepackage{longtable}

\usepackage{float}

\makeatletter
\AtBeginDocument{\@ifpackageloaded{rotating}{\PassOptionsToPackage{figuresright}{rotating}}{\usepackage[figuresright]{rotating}}\setlength{\rotFPtop}{0pt plus 1fil}\setlength{\rotFPbot}{0pt plus 1fil}}
\makeatother

\makeatletter
 \AtBeginDocument{%
  \@ifpackagewith{endfloat}{figuresonly}
  {\DeclareDelayedFloatFlavor{sidewaysfigure}{figure}}%true
  {\@ifpackagewith{endfloat}{tablesonly}{\DeclareDelayedFloatFlavor{sidewaystable}{table}\DeclareDelayedFloatFlavor{longtable}{table}\DeclareDelayedFloatFlavor{landscape}{table}}%true
  {\@ifpackageloaded{endfloat}{\DeclareDelayedFloatFlavor{sidewaysfigure}{figure}\DeclareDelayedFloatFlavor{sidewaystable}{table}\DeclareDelayedFloatFlavor{longtable}{table}\DeclareDelayedFloatFlavor{landscape}{table}}{}}%false
  }%false
  }
\makeatother

\usepackage{pdflscape}

\begin{document}

\begin{frontmatter}

\title{
    Dynamic prediction of time to event with survival curves    
}
    
\author[]{Jie Zhu\corref{c-d5abcdd03da0}}
\ead{elliott.zhu@unsw.edu.au}\cortext[c-d5abcdd03da0]{Corresponding author.}
\author[]{Blanca Gallego}
    
\address{Centre for Big Data Research in Health (CBDRH)\unskip, 
    University of New South Wales\unskip, Kensington\unskip, 2052\unskip, NSW\unskip, Australia}

\begin{abstract}
With the ever-growing complexity of primary health care system, proactive patient failure management is an effective way to enhancing the availability of health care resource. One key enabler is the dynamic prediction of time-to-event outcomes. Conventional explanatory statistical approach lacks the capability of making precise individual level prediction, while the data adaptive binary predictors does not provide nominal survival curves for biologically plausible survival analysis. The purpose of this article is to elucidate that the knowledge of explanatory survival analysis can significantly enhance the current black-box data adaptive prediction models. We apply our recently developed counterfactual dynamic survival model (CDSM) to static and longitudinal observational data and testify that the inflection point of its estimated individual survival curves provides reliable prediction of the patient failure time. 
\end{abstract}
\begin{keyword} 
Survival Analysis\sep Bayesian Machine Learning\sep Predictive Modelling\sep Data Mining
\end{keyword}

\end{frontmatter}
    
\section{Introduction}
Time-to-event (TTE) predictions are extensively used by medical statisticians. Traditional methods of logistic regression are not suited to include both the event and time aspects as the outcome in the model. Non-parametric models such as the Kaplan-Meier\unskip~\cite{904956:20541023} estimator and the semi-parametric Cox proportional hazard models and its extentsions\unskip~\cite{904956:20513280,904956:20513265} face the challenge of adjusting for multiple/time-varying covariates. The recent development of data adaptive models such as the deep neural networks\unskip~\cite{904956:20513279} and Super-Learner\unskip~\cite{904956:20541254} enable the efficient estimation of individual survival curves with static and longitudinal data, yet relatively little has been written about the implication of these explanatory techniques in the context of event time prediction.

The strength of explanatory survival analysis has been applied in data adaptive predictive models to improve the estimation accuracy of survival curves. In Deephit\unskip~\cite{904956:20513276}, a rank loss function is designed to evaluate whether the model can order observations  by their expected time to fail; in DeepSurv\unskip~\cite{904956:20543860}, authors approximates the Cox proportional hazard function using a densely connected neural network; and in WTTE-RNN\unskip~\cite{904956:20543171}, the predicted event time is assumed to follow a Weibull distribution whose parameters is estimated using a recurrent neural network. These model are in contrast with conventional binary predictors such as the recurrent neural networks proposed in the 2019 PhysioNet Challenge\unskip~\cite{904956:20513258}, where the prediction of TTE was equated to a longitudinal binary classification problem. 

In our recently proposed counterfactual dynamic survival model (CDSM), we relaxed major limitations of the three models aforementioned. Specifically, we do not assume Cox proportional hazard ratio or any parametric assumption in our model. At the same time, we allow longitudinal covariates and quantify the uncertainty of the neural network estimations using Bayesian dense layers. The focus of our previous work is the model development and its application to causal inference. In this study, we fixate on the prediction power of CDSM as an outcome model and explore how biologically plausible survival curve estimations can improve  the TTE predictions. 

In Section 2, we describe the methodology to estimate the survival outcomes and predict the time to event. Section 3 introduces a set of case studies and model evaluation techniques. Results are presented in section  4. We end our study with a discussion. 
    
\section{Predicting the time to event with survival curves }

To formalize the framework for causal inference for longitudinal survival outcomes, we follow the notations in previous study \citep{793296:18897670}. Suppose we observe a sample $\mathcal{O} \text{ of } n $ independent observations generated from an unknown distribution $\mathcal{P}_0 $ :
\begin{eqnarray*}\mathcal{O} := \big(X_i(t), Y_i(t), A_i(t), t_{i} = \min(t_{s,i},t_{c,i})\big),  i~= 1,2,\ldots, n  \end{eqnarray*}
where $X_i(t) = (X_{i,1}(t), X_{i,2}(t),\ldots,X_{i,\textit{d}}(t)), d=1,2, \ldots,D $ are baseline covariates at time $t $, $\;t=1,2,\dots,\Theta, $ with $\Theta $ being the maximum follow-up time of the study; $A(t)_i $ is the treatment condition at time $t $, $A(t)_i=1 $ if observation $i $ receives the treatment and $A(t)_i=0 $ if it is under control condition; $Y_i(t) $ denotes the outcome at time  $t $, $Y_i=1 $  if $i $ experienced an event and $Y_i=0
 $ otherwise; $t_{i} $ is determined by the event or censor time, $t_{s,i}$ or $t_{c,i}$, whichever happened first.

For each individual $i $, we define the hazard rate  $h(t)$, the probability of experiencing an event in interval $(t-1,t]$, as:
\begin{align}
\label{dfg-9cccfce24bde}
h(t)\;:=\;Pr(Y(t)=1\;\vert\;\overline A(t,u),\overline X(t,u)),
\end{align}

\noindent where $\bar{A}$ and $\bar{X}$ are the history of treatments and covariates from $t-u $ to $t-1 $ with $u $ being the length of the observation history. Thus, the probability that an uncensored individual will experience the event in time $t$ can be written as a product of terms, one per period, describing the conditional probabilities that the event did not occur since time $0$ to $t-1$ but occur in period $(t-1,t]$:

\begin{align*}
\begin{split}
Pr(t_{s,i} = t)&= h(t)(1-h(t-1))(1-h(t-2))\cdots(1-h(0)) \\
 						&= h(t) \prod_{j=0}^{t-1}(1-h(j)).
\end{split}
\end{align*}

\noindent similarly, the probability that a censored individual will experience an event after time $t$ can be written as a product of terms describing the conditional probabilities that the event did not occur in any observations:

\begin{align}
\begin{split}
\label{dfg-2c1cf113437b}
S(t) &= Pr(t_{s,i} > t) \\
&= (1-h(t))(1-h(t-1))(1-h(t-2))\cdots(1-h(0)) \\
&= \prod_{j=0}^{t}(1-h(j)).
\end{split}
\end{align}

\noindent which is the population survival function. 

The outcome label $Y^M$ for individual i is defined as a matrix over each time period $t\in[1,2,\ldots, \Theta]$: 

\begin{align}
\begin{split}
\label{dfg-364b5b929424}
Y^M_i&=\begin{bmatrix}E_i\\C_i\end{bmatrix},\\
\text{where }\\
E_i&= (e_i(t=1),e_i(t=2),\dots,e_i(t=t_i)  ,\ldots,e_i(t=\Theta))\\
e_i(\cdot) &= 1 \text{ for } t<t_i \text{ if } i \text{ is censored or having an event at  } t_i \\
e_i(\cdot) &= 0 \text{ for } t\geq t_i;\\
C_i&= (c_i(t=1),c_i(t=2),\ldots,c_i(t=t_i) ,\ldots,c_i(t=\Theta))\\
c_i(\cdot) &= 0 \text{ if } i \text{ is censored at } t_i; \\
c_i(\cdot) &= 0 \text{ and } c_i(t_i) = 1 \text{ if } i \text{ is having an event at } t_i.
\end{split}
\end{align}

\noindent The output of our model will be a $\Theta$-dimension vector, $\hat{H}_{i,\Theta}$, and each element represents the predicted conditional probability of surviving a time interval, which will be $\{1-\hat{h}_i(j)$, for $j\in[0,1,2,\ldots,\Theta]\}$. Then the estimated survival curve will be given by $\hat{Y}_i(t)=\prod_{j=0}^{\Theta}(1-\hat{h}_i(j))$.

\bgroup
\fixFloatSize{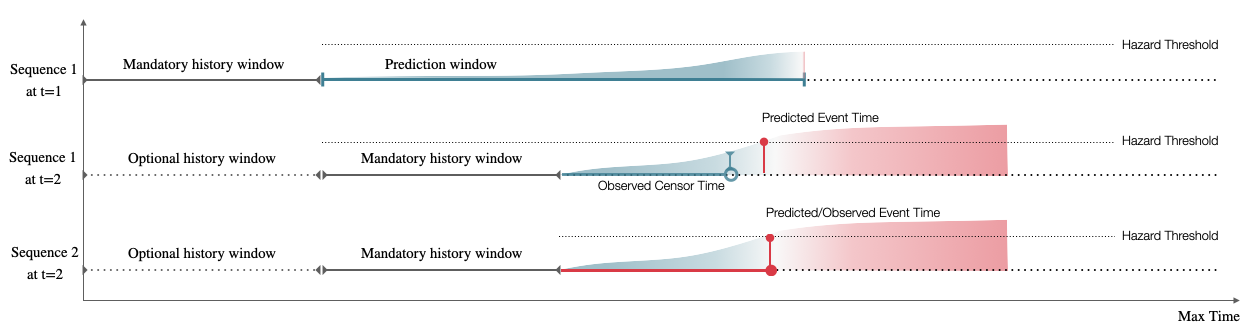}
\begin{figure*}[!htbp]
\centering \makeatletter\IfFileExists{images/screen-shot-2020-10-25-at-9-u54-u12-pm.png}{\includegraphics[width=.96\linewidth]{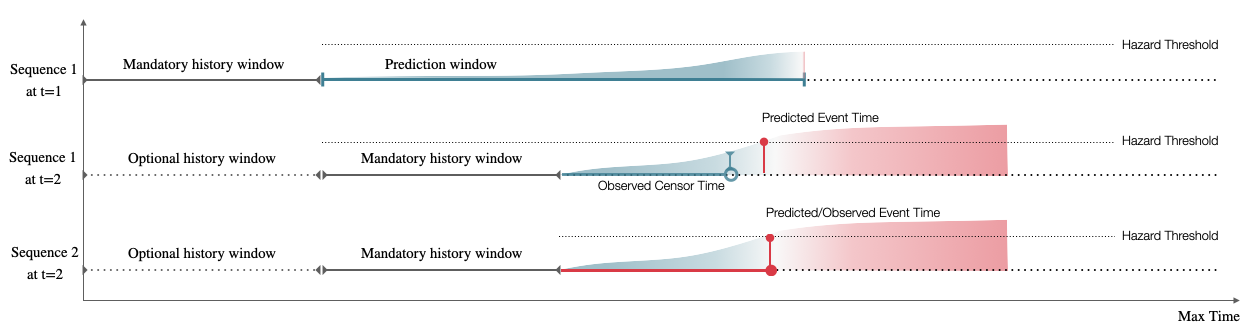}}{}
\makeatother 
\caption{{Prediction of event time based on hazard curves. }}
\label{f-c4c9fa151d11}
\end{figure*}
\egroup
Conventional predictive models fit the multivariate logistic outcome $\hat{Y}_{1\theta} $ in Equation~(\ref{dfg-364b5b929424}) using binary classifiers, where researchers have to set the optimal probability threshold to classify whether an event will occur (see hazard threshold in Figure~\ref{f-c4c9fa151d11}). For instance, one can use the Nelder-Mead method to locate the optimal probability threshold via minimizing the distance between the actual and predicted event time, yet this in-sample threshold might not be the optimal for predicting the TTE on a new cohort.

This study attempts to learn from the biological survival curve and uses the inflection point, $\hat{T}  $, of the survival curve in Equation~(\ref{dfg-2c1cf113437b}) to signify the event time, which we define as the time point equateing the second derivative of estimated survival curve $\hat{S} $ to zero:
\let\saveeqnno\theequation
\let\savefrac\frac
\def\dispfrac{\displaystyle\savefrac}
\begin{eqnarray}
\let\frac\dispfrac
\gdef\theequation{4}
\let\theHequation\theequation
\label{dfg-beabf65fc547}
\begin{array}{@{}l}\hat{T} = t \text{ where }  \frac{\partial^{2} \hat{S}}{\partial t^{2}} = 0\end{array}
\end{eqnarray}
\global\let\theequation\saveeqnno
\addtocounter{equation}{-1}\ignorespaces 
In Figure~\ref{f-c4c9fa151d11}, we can see the hazard rate has a rapid increase after  $\hat{T}  $, which means a high probability of experiencing an event. The uncertainty of the estimated survival curve quantifies the uncertainty of the predicted event time.
    
\section{Study design and databases}
We built and then validated a survival outcome model based on the retrospective analysis of three static databases and three dynamic longitudinal databases. The summary of these data sets are presented in Table~\ref{tw-fc13acc87430}.

\begin{table*}[!htbp]
\caption{{Summary of the clinical data sets. } }
\label{tw-fc13acc87430}
\def\arraystretch{1}
\ignorespaces 
\centering 
\begin{tabulary}{\linewidth}{LLLLL}
\tbltoprule Database & Sample & Covariates & Unique Time Points & \% Censored\\
\tblmidrule 
SUPPORT  &
  8873 &
  14 &
  1714 &
  32\%\\
METABRIC  &
  1904 &
  9 &
  1686 &
  42\%\\
GBSG  &
  2232 &
  7 &
  1230 &
  43\%\\
PhysioNet &
  40336 &
  40 &
  20*2 hours &
  93\%\\
MIMIC-III &
  20938 &
  44 &
  20*2 hours &
  86\%\\
CPRD AF &
  18102 &
  53 &
  20*3 months &
  82\%\\
\tblbottomrule 
\end{tabulary}\par 
\end{table*}
The static data sets were provided by the DeepSurv python package\unskip~\cite{904956:20543860} which includes:

\begin{enumerate}
\item \relax The Study to Understand Prognoses and Preferences for Outcomes and Risks of Treatments (\textbf{SUPPORT}); 
\item \relax The Molecular Taxonomy of Breast Cancer International Consortium (\textbf{METABRIC}); and 
\item \relax The Rotterdam tumor bank and German Breast Cancer Study Group (\textbf{GBSG}).
\end{enumerate}
The longitudinal data sets are:

\begin{enumerate}
\item \relax The Medical Information Mart for Intensive Care version III (\textbf{MIMIC-III}), an open-access, anonymized database of 61,532 admissions from 2001{\textendash}2012 in six ICUs at a Boston teaching hospital\unskip~\cite{904956:20513259} .
\item \relax 2019 PhysioNet Sepsis prediction challenge data set\unskip~\cite{904956:20513258}  (\textbf{PhysioNet}).  PhysioNet Sepsis prediction c containing more than 3.3 million admissions from 2003{\textendash}2016 in 459 ICUs across the United States.
\item \relax The Clinical Practice Research DataLink data set\unskip~\cite{904956:20513256}  (\textbf{CPRD AF}) comparing Vitamin K Antagonists (VKAs) and Non-Vitamin K antagonist oral anticoagulants (NOAC) in preventing three combined outcomes (ischemic attack, major bleeding and death) of patients with non-valvular atrial fibrillation (AF).
\end{enumerate}
In both MIMIC-III and PhysioNet, we define Sepsis event as a suspected infection (prescription of antibiotics and sampling of bodily fluids for microbiological culture) combined with evidence of organ dysfunction, defined by a two points deterioration of SOFA score\unskip~\cite{904956:20513252}. We follow previous papers\unskip~\cite{904956:20513258,904956:20513251}  for data extraction and processing.  For the PhysioNet, we combined data from hospital A and B, and used hospital location (A or B) as the synthetic treatment condition. For MIMIC-III, we define the treatment as the usage of mechanical ventilation(MV). For the CPRD AF, the outcome of interest is the first occurrence of combined outcomes of major bleeding, death and stroke. The treatment is the usage of NOAC vs the control of using VKAs. 

For the static data sets, we discretized the time points into windows of 50 time steps and censored all steps do not form a complete window (i.e. $34 \equiv 1714 \mod 50 $ windows for the SUPPORT data set, $33 \equiv 1686 \mod 50 $ windows for  METABRIC and $24 \equiv 1230 \mod 50 $ windows for GBSG). For the longitudinal data sets, we considered the first 20 time stamps for each patient (i.e., the first 20 2-hour intervals for PhysioNet and MIMIC-III, and the first 20 months for the AF.). We split each database into estimation data set (70\% of the original data for training and 10\% validation) and testing data set (20\% of the original).

\subsection{Model evaluation}We performed an evaluation of the estimations of survival curves and predictions of event time using the three metrics described below:

\textbf{The area under the receiver operating characteristic (AUROC) and \textbf{C-Index: }} we use AUROC and Harrell's C-index\unskip~\cite{904956:20513255}  to evaluate the models' discrimination performance. Both indicators are calculated using the multivariate logistic outcome $\hat{Y}_{1\theta} $ in Equation~(\ref{dfg-364b5b929424}) .

\textbf{Utility distance (Distance Score): }we define the distance metric to evaluate the predicted event time as: 
\begin{eqnarray*}\text{Distance} = \frac{1}{n}\sum_{i} |\hat{T}_i - T_i| \end{eqnarray*}
where $\hat{T}  $ is defined in Equation~(\ref{dfg-beabf65fc547})  and $T $ is the true event/censoring time. 

We compared following algorithms on the estimation of survival curves and the prediction of event time:

\begin{itemize}
  \item \relax Dynamic Bayesian survival causal model (\textbf{CDSM}): the model targets the outcome defined in Equation~(\ref{dfg-364b5b929424}) by training two counterfactual sub-networks for treated and controlled observations. If no treatment variable is defined, we create two copies of the original data set, with first one marked as receiving the treatment and the second one as under control. The loss function of CDSM has three components: 1) the partial log likelihood loss of the joint distribution of the first hitting time and corresponding event or right-censoring; 2) the rank loss function to capture the concordance score defined in survival analysis; 3) The calibration loss function minimizes the selection bias in for treatment assignment. Please refer to our previous paper for details. 
  \item \relax Plain recurrent neural network with survival outcomes (\textbf{RNN}): the model modifies the CDSM by removing the counterfactual sub-networks and the third loss function in CDSM. No treatment variable has to be specified in this model. 
  \item \relax Plain recurrent neural network with binary outcomes (\textbf{RNN Binary}): the model provides the direct prediction on the longitudinal outcome $\hat{Y}_{1\theta} $ in Equation~(\ref{dfg-364b5b929424}) using the mean squared error loss function. 
  \item \relax DeepHit\unskip~\cite{904956:20550472}: the model uses the same loss functions as the RNN but does not capture the history of covariates and is only evaluated for static databases.
\end{itemize}
  The model construction and training uses Python 3.8.0 with Tensorflow 2.3.0 and Tensorflow-Probability 0.11.0\unskip~\cite{904956:20513254} (code available at \BreakURLText{https://github.com/EliotZhu/CDSM).}

\section{Results}
In Table~\ref{tw-e8b3430b700e}, we confirmed CDSM, RNN and DeepHit had similar performance on the estimation of survival curves (see the concordance index) and the prediction of event time (see the distance score) in the three static testing data sets. However, in terms of the AUROC, we noticed RNN Binary had superior performance than the others, although it had lower C-Index. 

The counterfactual sub-networks and the selection bias calibration loss function in CDSM did not affect the estimation accuracy, resulting the equivalency among CDSM, RNN and DeepHit in the static non-causal survival estimations.

\begin{table}[!htbp]
\caption{{Model performance on static datasets} }
\label{tw-e8b3430b700e}
\centering 
\begin{threeparttable}

\def\arraystretch{1}
\ignorespaces 
\centering 
\begin{tabulary}{\linewidth}{p{\dimexpr.20\linewidth-2\tabcolsep}p{\dimexpr.20\linewidth-2\tabcolsep}p{\dimexpr.1917\linewidth-2\tabcolsep}p{\dimexpr.2083\linewidth-2\tabcolsep}p{\dimexpr.20\linewidth-2\tabcolsep}}
\tbltoprule 
\textbf{Dataset } &
  \textbf{Metabric } &
   &
   &
  \\
\textbf{Metrics } &
  \textbf{CDSM} &
  \textbf{RNN} &
  \textbf{RNN Binary } &
  \textbf{DeepHit }\\
AUROC  &
  0.869 &
  0.877 &
  \textbf{0.885 } &
  0.874\\
C-Index &
  \textbf{0.685 } &
  0.655 &
  0.590 &
  0.683\\
Distance Score &
  4.186 &
  4.034 &
  \textbf{3.944 } &
  4.097\\\cline{1-1}\cline{2-2}\cline{3-3}\cline{4-4}\cline{5-5}
 &
  \textbf{GBSG} &
   &
   &
  \\
AUROC  &
  0.781 &
  0.780 &
  \textbf{0.817 } &
  0.798\\
C-Index &
  \textbf{0.617 } &
  0.593 &
  0.559 &
  0.613\\
Distance Score &
  \textbf{4.384} &
  4.974 &
  4.464 &
  4.668\\\cline{1-1}\cline{2-2}\cline{3-3}\cline{4-4}\cline{5-5}
 &
  \textbf{Support} &
   &
   &
  \\
AUROC  &
  0.792 &
  0.788 &
  \textbf{0.802} &
  0.820 \\
C-Index &
  \textbf{0.653} &
  0.650 &
  0.550 &
  0.633\\
Distance Score &
  \textbf{4.434} &
  5.474 &
  5.672 &
  4.231 \\
\tblbottomrule 
\end{tabulary}\par 
\begin{tablenotes}\footnotesize 
    
\item{All metrics are averaged over estimation windows using testing data sets. The best value in each metric is in bold. }
\end{tablenotes}
\end{threeparttable}

\end{table}
Similar trend was observed when we evaluated CDSM, RNN, and RNN Binary using the longitudinal databases (see the estimation data set evaluations in Table~\ref{tw-03a37ae9b5b0}). However,  in the corresponding testing data sets, CDSM significantly outperformed RNN Binary, especially for the C-Index and distance score. We saw the imposition of survival outcome in Equation~(\ref{dfg-364b5b929424})  and concordance loss functions defined in CDSM/RNN produced nominal survival curves, where the RNN Binary only maximized the discrimination performance on the binary indicator of whether the sepsis has occurred (i.e., the estimated survival probabilities for the AF testing data set were stacked at zeros and ones as shown in Figure~\ref{f-7d50cdc051bf} (a)).

\begin{landscape}
\makeatletter\@twocolumnfalse\makeatother
\begin{ThreePartTable}

\centering 
\begingroup
\makeatletter\if@twocolumn\@ifundefined{theposttbl}{\gdef\TwoColDocument{true}\onecolumn\onecolumn}{}\fi\makeatother 
\begin{TableNotes}\footnotesize 
    
\item{All metrics are averaged over 20 estimation windows using either estimation (the default) or testing data sets (specified in brackets). The best value in each metric is in bold.}
\end{TableNotes}\setlength\LTcapwidth{\textheight}
\begin{longtable}{p{\dimexpr.10000000000000004\linewidth-2\tabcolsep}p{\dimexpr.10000000000000004\linewidth-2\tabcolsep}p{\dimexpr.10000000000000004\linewidth-2\tabcolsep}p{\dimexpr.10000000000000004\linewidth-2\tabcolsep}p{\dimexpr.10000000000000004\linewidth-2\tabcolsep}p{\dimexpr.10\linewidth-2\tabcolsep}p{\dimexpr.10\linewidth-2\tabcolsep}p{\dimexpr.10\linewidth-2\tabcolsep}p{\dimexpr.10\linewidth-2\tabcolsep}p{\dimexpr.10\linewidth-2\tabcolsep}}
\caption{{Model performance on dynamic datasets} }
\label{tw-03a37ae9b5b0}
\def\arraystretch{1}\\\endfirsthead \hline \noalign{\vskip3pt} \noalign{\textit{Table \thetable\ continued}} \noalign{\vskip3pt} 
\tbltoprule \textbf{Dataset} & \multicolumn{3}{p{\dimexpr(.3000000000000001\linewidth-2\tabcolsep)}}{\textbf{PhysioNet (hours)}} & \multicolumn{3}{p{\dimexpr(.30000000000000004\linewidth-2\tabcolsep)}}{\textbf{MIMIC-III (hours)}} & \multicolumn{3}{p{\dimexpr(.30\linewidth-2\tabcolsep)}}{AF \textbf{(months)}}\\
\tblmidrule \endhead \hline \noalign{\vskip3pt} \noalign{\textit{\hfill Continued on next page}} \noalign{\vskip3pt} \endfoot \insertTableNotes \endlastfoot 
\tbltoprule \textbf{Dataset} & \multicolumn{3}{p{\dimexpr(.3000000000000001\linewidth-2\tabcolsep)}}{\textbf{PhysioNet (hours)}} & \multicolumn{3}{p{\dimexpr(.30000000000000004\linewidth-2\tabcolsep)}}{\textbf{MIMIC-III (hours)}} & \multicolumn{3}{p{\dimexpr(.30\linewidth-2\tabcolsep)}}{AF \textbf{(months)}}\\
\tblmidrule 
\textbf{Metrics} &
  \textbf{CDSM} &
  \textbf{RNN} &
  \textbf{RNN Binary} &
  \textbf{CDSM} &
  \textbf{RNN} &
  \textbf{RNN Binary} &
  \textbf{CDSM} &
  \textbf{RNN} &
  \textbf{RNN Binary}\\
AUROC &
  0.980 &
  0.984 &
  \textbf{0.997} &
  0.959 &
  0.953 &
  \textbf{0.988} &
  0.978 &
  0.986 &
  \textbf{0.995}\\
AUROC (test) &
  \textbf{0.869} &
  0.858 &
  0.824 &
  0.969 &
  0.941 &
  \textbf{0.983} &
  \textbf{0.984} &
  0.983 &
  0.933\\
C-Index &
  0.991 &
  0.985 &
  \textbf{0.992} &
  0.749 &
  \textbf{0.851} &
  0.823 &
  0.871 &
  0.880 &
  \textbf{0.885}\\
C-Index (test) &
  \textbf{0.874} &
  0.837 &
  0.776 &
  \textbf{0.751} &
  0.653 &
  0.682 &
  \textbf{0.877} &
  0.863 &
  0.785\\
Distance Score &
  3.388 &
  4.034 &
  \textbf{2.017} &
  2.230 &
  2.410 &
  \textbf{2.191} &
  1.331 &
  1.082 &
  \textbf{0.751}\\
Distance Score (test) &
  \textbf{3.047} &
  3.635 &
  3.767 &
  \textbf{2.291} &
  2.375 &
  2.339 &
  1.116 &
  \textbf{1.035} &
  1.240\\
Score Std &
  \textbf{11.586} &
  7.584 &
  0.248 &
  \textbf{1.793} &
  0.498 &
  0.041 &
  \textbf{3.294} &
  1.753 &
  0.009\\
Score Std (test) &
  \textbf{11.522} &
  6.813 &
  0.080 &
  \textbf{1.743} &
  0.557 &
  0.038 &
  \textbf{3.118} &
  1.469 &
  0.011\\
\tblbottomrule 
\end{longtable}
\endgroup
\end{ThreePartTable}

\makeatletter\@ifundefined{TwoColDocument}{}{\twocolumn}\makeatother 
\end{landscape}

\bgroup
\fixFloatSize{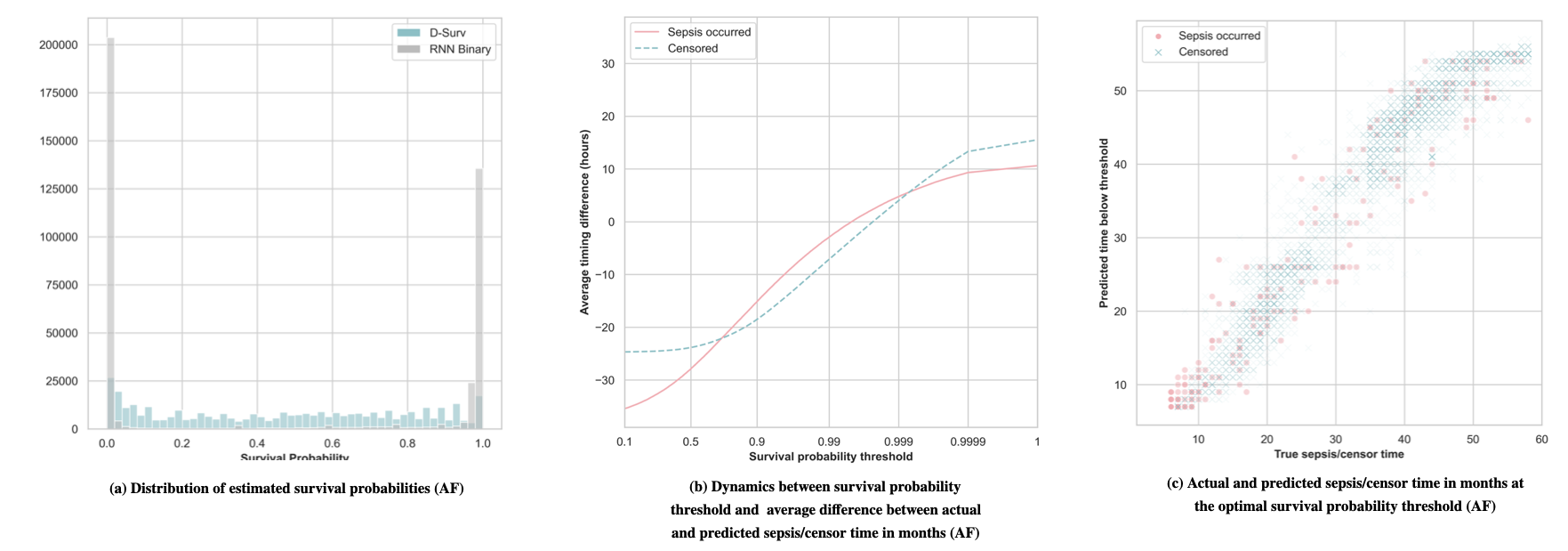}
\begin{figure*}[!htbp]
\centering \makeatletter\IfFileExists{images/screen-shot-2020-10-28-at-10-u40-u34-pm.png}{\includegraphics[width=.98\linewidth]{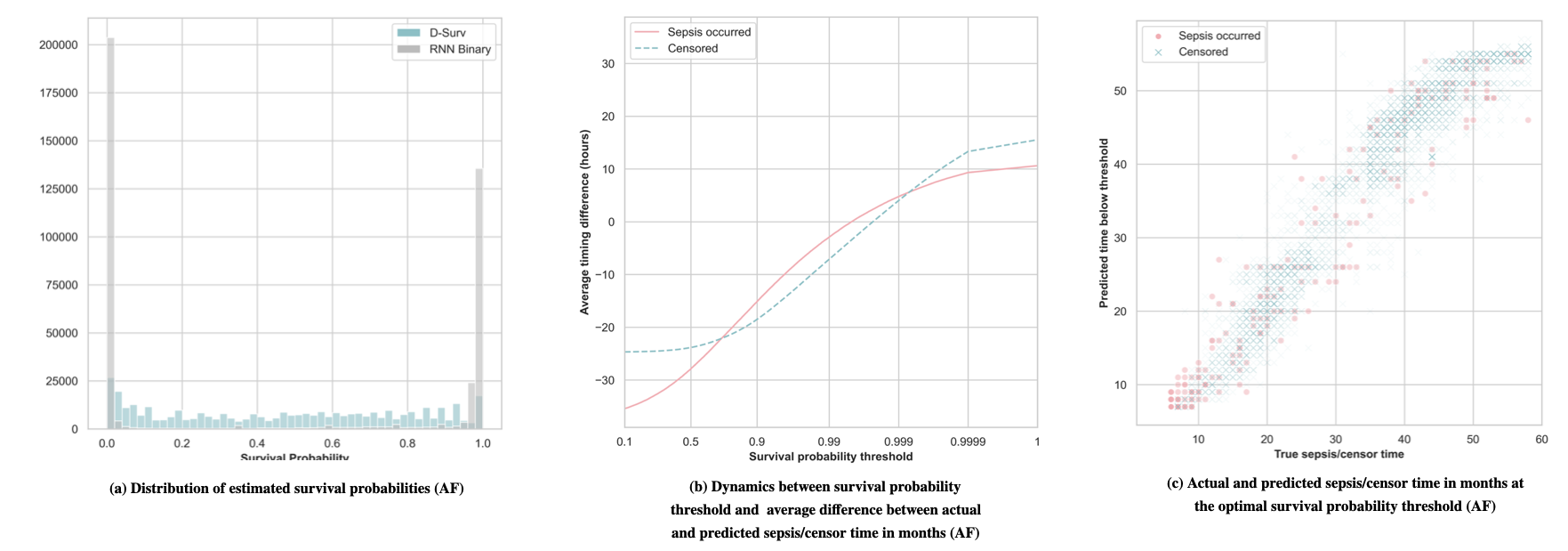}}{}
\makeatother 
\caption{{\textbf{Diagnostic plots for event time prediction with AF testing data set }} (a) distribution of estimated survival probabilities across all time points for AF testing data set by benchmark algorithms; (b) average difference between the predicted and true event/censor time estimated using probability threshold approach for AF testing data set; and (c) scatter plot of predicted and true event/censor time estimated using the inflection point approach for AF testing data set.}
\label{f-7d50cdc051bf}
\end{figure*}
\egroup
The nominal survival curves by CDSM made it possible to apply Equation~(\ref{dfg-beabf65fc547})  to locate the inflection point as the event time. This is a better approach than choosing a probability threshold to construct a Binary classifier. In Figure~\ref{f-7d50cdc051bf} (b), we saw the error of predicted event time is sensitive to the chosen probability threshold, where the range of average timing difference was from -4.3 to 9.5 months in a small threshold range: 0.99 to 0.9999. In contrast, after applying  the inflection point to determine the event time, we observed the predicted time accurately tracked the true time in Figure~\ref{f-7d50cdc051bf} (c), with most predictions happened ahead of the true AF event time. The average distance to from the predicted time is 1.720 months ahead of the true AF event time, while 1.014 months ahead for true AF censored time.  CDSM allows the threshold-free prediction of the individual event time and early intervention on patients who might be prone to event occurrence. 
    
\section{Discussion}
This study demonstrated that injecting the knowledge of survival analysis into the design of recurrent neural network can significantly improve the prediction of time-to-event outcomes. Our proposed outcome model, CDSM fitting the joint distribution of both failure and censored observations. The conventional machine learning algorithms for binary discrimination can maximize evaluation scores such as AUROC, but failed to provide meaningful survival curves and reliable predictions of event time. The major drawback of these algorithms, as identified by our empirical study, is that they do not take account of censoring and had significant drop in accuracy when being evaluated on the testing database. 

Among these applications, few has discussed the use of statistical survival analysis to infer the event time. Rather, straightforward binary classification algorithms are used to hard code the inference questions into binary states at each time stamp. Researchers then apply a probability threshold to determine whether the event will happen at a point in time.

\section*{Acknowledgments}This work was supported by National Health and Medical Research Council, project grant no. 1125414.

\bibliographystyle{elsarticle-num}
\bibliography{article.bib}

\end{document}